\documentclass{ecai}
\usepackage{times}
\usepackage{graphicx}
\usepackage{latexsym}
\usepackage{blkarray}
\usepackage{multirow}
\usepackage{subcaption}
\usepackage{graphics}
\usepackage{bm}
\usepackage{amsthm, amsmath, amssymb}
\usepackage{url}
\usepackage{algorithm}
\usepackage[noend]{algpseudocode}
\usepackage{amssymb} 
\usepackage{xpatch}

\makeatletter
\xpatchcmd{\@thm}{.}{}{}{}
\makeatother

\theoremstyle{definition}
\newtheorem{defi}{Definition}

\theoremstyle{plain}

\theoremstyle{remark}

\DeclareMathOperator*{\argmax}{arg\,max}

\begin{document}

\title{TIMELY: Improving Labeling Consistency in Medical\\
Imaging for Cell Type Classification}

\author{Yushan Liu\institute{Technical University of Munich and Siemens AG, Germany, \newline e-mail: yushan.liu@tum.de} , 
Markus M.\,Geipel\institute{Siemens AG, Germany, e-mail: markus.geipel@siemens.com} , Christoph Tietz\institute{Siemens AG, Germany, e-mail: christoph.tietz@siemens.com} , 
Florian Buettner\institute{Siemens AG, Germany, e-mail: buettner.florian@siemens.com}}

\maketitle
\bibliographystyle{ecai}

\begin{abstract}
Diagnosing diseases such as leukemia or anemia requires reliable counts of blood cells. Hematologists usually label and count microscopy images of blood cells manually. In many cases, however, cells in different maturity states are difficult to distinguish, and in combination with image noise and subjectivity, humans are prone to make labeling mistakes. This results in labels that are often not reproducible, which can directly affect the diagnoses. We introduce TIMELY, a probabilistic model that combines pseudotime inference methods with inhomogeneous hidden Markov trees, which addresses this challenge of label inconsistency. 
We show first on simulation data that TIMELY is able to identify and correct wrong labels with higher precision and recall than baseline methods for labeling correction. We then apply our method to two real-world datasets of blood cell data and show that TIMELY successfully finds inconsistent labels, thereby improving the quality of human-generated labels. 
\end{abstract}

\section{INTRODUCTION}
Manually labeling and counting cells based on morphology is an essential component in diagnosing blood diseases such as leukemia or anemia. Consequently, the quality of labels directly affects diagnoses and patient outcomes. In many cases, however, cell types from the same development line are difficult to distinguish, and in combination with image noise, lack of concentration, and subjectivity, humans are prone to make labeling mistakes. The labels are often inconsistent and not reproducible, and for example, in an experiment about classifying lymphocytes, "$31\%$ of the morphologists were not able to reproduce their previous classification" \cite{vanderMeer.2007}.

Existing approaches to improve labeling reproducibility and consistency focus on automation via machine learning-based classification, such as decision trees, support vector machines, or neural networks \cite{Akrimi.2014, Habibzadeh.2018, Kihm.2018, Shpilman.2017, Tomari.2014, Wheeless.1994}. However, training these algorithms relies on manually generated, noisy labels, and propagating this noise through training may result in biased predictions. To mitigate the adverse effect of label noise on predictive power, different approaches exist \cite{Frenay.2014}: Algorithms can be applied that are relatively robust to label noise \cite{Abellan.2010, Ratsch.2003, Sastry.2010}, filtering methods can be used to remove mislabeled instances from the training set \cite{Sanchez.1997b, Thongkam.2008, Wilson.2000}, or label noise can be modeled explicitly \cite{Bouveyron.2009, Lawrence.2001, Rantalainen.2011}. For example, the algorithms $k$-nearest neighbors ($k$-NN) and $k$-nearest centroid neighbors ($k$-NCN) \cite{Sanchez.1997} find instances whose classes do not agree with the classes of their neighbors \cite{Saez.2015, Sanchez.2003}. Alternative approaches find labeling errors based on confident learning \cite{cleanlab, Northcutt.2019, Northcutt.2017}, which estimates the joint distribution between noisy labels and true labels to identify noisy labels and improve training. While these algorithms are able to find noisy or wrong labels, they do not provide suggestions on what the correct labels should be. In addition, they are based on the premise of complete automation, which, especially in the medical domain, is often not feasible due to regulatory constraints requiring human oversight and accountability.

In this work, we follow an orthogonal, human-centered approach: Rather than taking the human (the hematologist performing the labeling task) out of the loop, we develop a human-centered interpretable AI algorithm that comprises two stages. In the first stage, our algorithm identifies those labels that are inconsistent with the morphology of the cells in an unsupervised manner, i.\,e., without training on noisy labels. In the second stage, an alternative, consistent label is suggested to the hematologist performing the labeling task, based on the labels of the cells in the direct neighborhood of the cell in question. We ensure interpretability of the method by explicitly incorporating prior knowledge on the biological system and an expert-driven error model into the state space model.

\begin{figure*}[h]
\begin{center}
\includegraphics[width=\linewidth]{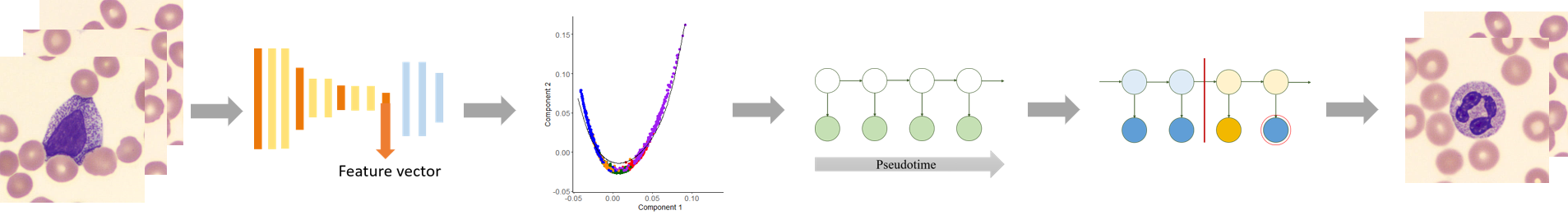} 
\end{center}
\caption{Overview of TIMELY. Given cell images and noisy labels, we use a neural network to learn meaningful feature vectors. Using a pseudotime inference algorithm, we order these vectors along a trajectory and estimate the pseudotimes. The corresponding labels are sorted accordingly and serve as the observations in a hidden Markov tree. The parameters of the hidden Markov tree are learned, and the hidden labels and transition borders are inferred. Last, we retrieve the images with inconsistent labels along with the proposed labels from the model.}
\label{fig:algorithm}
\end{figure*}

More specifically, pseudo\underline{t}ime \underline{i}nference with \underline{M}arkov mod\underline{e}ls for \underline{l}abeling consistenc\underline{y} (TIMELY) combines pseudotime inference algorithms with inhomogenenous hidden Markov trees, which is an extension of hidden Markov models.
Hidden Markov trees are used since blood stem and progenitor cells differentiate into more mature cell types during their development process, where blood cell lineages can branch into functionally distinct lineages. The differentiation process itself is commonly described as a stochastic process following the Markov assumption, where cells can either remain in the present type or differentiate into a child cell type \cite{abkowitz1996evidence}. Hidden Markov trees reflect this differentiation process and can be used to model the true, unobservable cell types (i.e. true labels) together with the noisy, observed expert labels.
Given a set of cell images and noisy labels, we establish an intrinsic ordering of the cells based on pseudotime inference methods. The ordered cells are used as input for the hidden Markov tree. 
We extend standard hidden Markov trees to the inhomogeneous case, propose parametric transition matrices and derive an efficient inference scheme. Finally, we identify inconsistent labels and propose alternative, consistent labels to the practitioner. An overview of TIMELY is shown in Figure~\ref{fig:algorithm}.

The manuscript is structured as follows. In Section $2$, we first describe how we establish an intrinsic ordering of the cells using pseudotime inference algorithms. We then briefly review hidden Markov trees and introduce an inhomogeneous extension with parametric transition matrices. Our method TIMELY, which combines both, is then described in detail. In Section $3$, we demonstrate, based on extensive simulations, that our modeling approach is able to identify and correct noisy labels with higher precision and recall than state-of-the-art methods for identifying noisy labels. Finally, we apply our algorithm to two real-world datasets of white blood cells with noisy labels in Section $4$ and demonstrate that our method is able to identify inconsistent labels and suggest alternative, consistent labels to practitioners in an interpretable manner. We validate these suggested new labels by reclassification via a domain expert.

\section{METHODS}

\subsection{Pseudotime inference}
The pseudotime of a cell describes the developmental progress of the cell along a dynamic process such as cell differentiation. The greater the pseudotime of a cell, the more mature is the cell. By using pseudotime inference algorithms, we can create a pseudotemporal ordering for all cells in a population. Pseudotime inference algorithms are usually applied on single-cell gene expression similarity measurements \cite{Haghverdi.2016}, where adjacent cells have higher expression similarity. We can apply these algorithms to medical images by interpreting the pixels of a cell image as information about the cell, similar to gene expression data, to obtain an ordering of the cells along trajectories.

There is a multitude of pseudotime inference methods to date, which differ in the requirement of existing prior information, scalability, and type of topology \cite{Saelens.2019}.
Most pseudotime inference methods consist of two parts. The first part is the calculation of a low-dimensional representation from the given expression data of the cells, and the second part is the ordering of the cells along an inferred trajectory.

Here, we use the algorithms SCORPIUS \cite{Cannoodt.2016} and STREAM \cite{Chen.2019}. SCORPIUS shows very good performance for linear datasets \cite{Saelens.2019}, while STREAM is well-suited for datasets with tree-like topologies.

In brief, given the expression profiles of the cells, SCORPIUS obtains a low-dimensional representation using multi-dimensional scaling (MDS). Next, SCORPIUS applies $k$-means clustering and sets the initial trajectory by connecting the cluster centers. The final trajectory results from an iterative refinement through the principal curves algorithm \cite{Hastie.1989}. The pseudotime is calculated by projecting the low-dimensional representations onto the trajectory. 

Similarly, STREAM first determines relevant features and then performs dimensionality reduction using modified locally linear embedding (MLLE). In the new embedding, an implementation of elastic principal graphs (ElPiGraph) \cite{Albergante.2018} is used to infer the trajectory and branching points. ElPiGraph approximates datasets with complex topologies by minimizing the elastic energy of the embedding and applying graph transformations. The cells are then projected onto the resulting tree according to their pseudotimes and their assigned branches.

\subsection{Hidden Markov trees}
Hidden Markov trees are used to describe the differentiation process of the cells, which is a stochastic process following the Markov assumption \cite{abkowitz1996evidence}. There is one root cell type, and all other cell types develop from it and can be mapped onto a tree-like topology reflecting their respective progeny. Assume that we know the topology of the dataset, i.\,e., we know the shape of the Markov tree.

\begin{defi}
A tree $\bar{Z}_1$ is a \textit{Markov tree} if for each leaf, the directed path connecting the root and the leaf is a Markov chain.
\end{defi}

A hidden Markov tree is an extension of a Markov tree, and it is used for applications where the Markov property does not hold or where the states can only be observed indirectly. The model consists of observed variables and hidden variables, where only the hidden variables follow the Markov property. The present observed variable depends on the present hidden state, but neither on previous observed states nor on previous hidden states. 

Define
$\bar{Z}_1 := (Z_1, \dots, Z_T)$ and $\bar{X}_1 := (X_1, \dots, X_T)$ for $T \in \mathbb{N}$
to be the hidden tree and the observed tree, respectively.
The roots of the trees are $Z_1$ and $X_1$, and both trees have the same indexing structure.

\begin{defi}
Let $\bar{X}_1$ and $\bar{Z}_1$ be two trees, where $\bar{X}_1$ is the observed tree and $\bar{Z}_1$ is the hidden tree.
The pair $(\bar{Z}_1, \bar{X}_1)$ is a \textit{hidden Markov tree} (HMT) if
\begin{itemize}
\item $\bar{Z}_1$ is a Markov tree, and 
\item the distribution of the observed variable $X_t$ depends only on the hidden variable $Z_t$ for all $t \in \{1, \dots, T\}$.
\end{itemize}
\end{defi}

For the application on cell image labels, the variable $X_t$ corresponds to the noisy (observed) expert label, and $Z_t$ represents the true (unobservable) label of the image, which may be different from the expert label. The sequence of images is sorted by increasing pseudotime, which has been calculated before by a suitable pseudotime inference algorithm. 
Let $K$ be the number of cell types and $T$ be the number of images in the dataset. 

\begin{defi}
The hidden Markov tree $(\bar{Z}_1, \bar{X}_1)$ is governed by the parameters $\bm{\pi} \in [0,1]^K$, $\bm{A}^{(t)} \in [0,1]^{K\times K}$, and $\bm{B} \in [0,1]^{K \times K}$. For $2 \leq t \leq T$, $1 \leq k, \Tilde{k} \leq K$, we define
\begin{align}
\bm{\pi}_k &:= \mathbb{P}(Z_1 = k),\\
\bm{A}_{kl}^{(t)} &:= \mathbb{P}(Z_{t} = l \mid Z_{\mathfrak{p}(t)} = k),\\
\bm{B}_{k \Tilde{k}} &:= \mathbb{P}(X_t = \Tilde{k} \mid Z_t = k),
\end{align}
where $\mathfrak{p}(t)$ denotes the parent of node $t$. We call $\bm{\pi}$ the \textit{start probabilities}, $\bm{A}^{(t)}$ the \textit{transition matrix} at node $t$, and $\bm{B}$ the \textit{emission matrix}. 
If the transition matrix $\bm{A}^{(t)}$ is independent of $t$, the model is called \textit{homogeneous}; otherwise, the model is called \textit{inhomogeneous}.
\end{defi}

The transition matrix $\bm{A}^{(t)}$ describes the probability of staying in the present cell type or changing to a child cell type.
The emission matrix $\bm{B}$ represents the expert labeling error model, where $\bm{B}_{k\Tilde{k}}$ is the probability that the expert predicts label $\Tilde{k}$ when the true cell type of the cell in the image is $k$.

\begin{figure}[h]
\begin{center}
\includegraphics[width=0.93\linewidth]{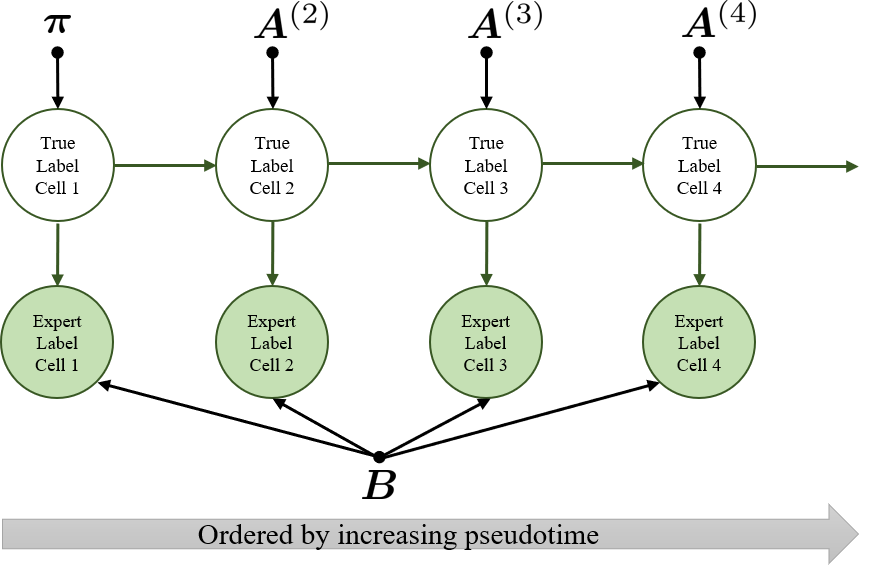} 
\end{center}
\caption{Hidden Markov model for our application. A hidden Markov model can be visualized as a graph, where the nodes represent the random variables and the edges the transitions and emissions. The expert labels serve as the observations and are ordered by pseudotime. The task is to find the true cell types.}
\label{fig:hmm_model}
\end{figure}

A hidden Markov model (HMM) is a special case of an HMT, where the underlying topology is a chain.
Figure \ref{fig:hmm_model} shows a visualization of an HMM for our application.

\paragraph{Time-dependent transition matrices} 
We use the following information to set up the parametric transition matrices. We know the topology of the dataset, and following the Markov assumption of blood cell differentiation \cite{abkowitz1996evidence}, it is only possible for a cell to stay in the same cell type or to transition to one of the child cell types (see Figure \ref{fig:dataset_tree} for a topology example). There is no way to skip one cell type or to go back to a previous cell type. Once one of the end stages is reached, there are no transitions anymore.

Standard homogeneous HMMs/HMTs are based on the assumption that the transition between states is independent of $t$, which would correspond to cells sampled uniformly across the development trajectory. However, in practice, this sampling (i.\,e., the labeled cells) are from arbitrary points on the development trajectory, which is reflected by large variation in pseudotime difference between neighboring cells. This difference directly affects the probability of a cell to transition to a different cell type. The larger the pseudotime difference between two cells, the greater is the likelihood for a transition (and the lower is the likelihood to remain in the same cell type). 
Consequently, the entries of the transition matrix at node $t$ should not only depend on the cell type of the previous cell, but also on the pseudotime difference between the present cell and the previous cell. To  model the dependency of the transition matrix on the pseudotime, we extend the algorithms for HMMs and HMTs to the inhomogeneous case and derive appropriate parametric transition matrices. 

Define $y_t \in \mathbb{R}_{\geq 0}$ as the pseudotime difference between node $t-1$ and node $t$, after they have been ordered by increasing pseudotime. To find reasonable entries for the transition matrices, we rewrite the transition probabilities at node $t$:

{
\begin{align}
\bm{A}_{kl}^{(t)} &= \mathbb{P}(Z_{t}=l | Z_{t-1}=k, y_t)\\[2mm]
&= \frac{\mathbb{P}(Z_{t}=l, Z_{t-1}=k, y_t)}{\mathbb{P}(Z_{t-1}=k, y_t)}\\[2mm]
&= \frac{\mathbb{P}(Z_{t-1}=k)\mathbb{P}(Z_{t}=l | Z_{t-1}=k)\mathbb{P}(y_t | Z_{t}=l, Z_{t-1}=k)}{\mathbb{P}(Z_{t-1}=k)\mathbb{P}(y_t | Z_{t-1}=k)}\\[2mm]
&= \frac{\mathbb{P}(Z_{t}=l | Z_{t-1}=k)\mathbb{P}(y_t | Z_{t}=l, Z_{t-1}=k) }{\mathbb{P}(y_t | Z_{t-1}=k)}.
\end{align}}

Define $\mathbb{P}(Z_{t}=l \mid Z_{t-1}=k) =: p_{kl} \in [0,1]$ to be the transition probability from cell type $k$ to cell type $l$. Let $p_{kl}$ be a constant independent of $t$, with condition $\sum_{l=1}^K p_{kl} = 1$ for all $k$. 

For the probability $\mathbb{P}(y_t \mid Z_{t}=l, Z_{t-1}=k)$, we know that the support of $y_t$ is $[0, \infty)$. Since we have no more information about the distribution of the pseudotime difference, we use the maximum entropy probability distribution. The least informative distribution for a random variable with support $[0, \infty)$ and mean $1/\lambda$ is the exponential distribution with rate $\lambda$. Let the rate $\lambda$ be dependent on the cell types $k$ and $l$. 

Then, for each possible transition in the cell lineage tree, the entry in the transition matrix after normalization has the form
\begin{equation}
\bm{A}_{kl}^{(t)} = \frac{p_{kl} \cdot\lambda_{kl} \exp(-\lambda_{kl} \cdot y_t)}{\sum_{i=1}^K p_{ki} \cdot \lambda_{ki}\exp(-\lambda_{ki} \cdot y_t)}
\label{eqn:transition_matrix_hmt}
\end{equation}
for $p_{ki} \in [0,1]$ and $\lambda_{ki} > 0$.

The parameters in \eqref{eqn:transition_matrix_hmt} are learned using the generalized EM algorithm \cite{Neal.1998} since the corresponding objective function is intractable. 
The generalized Viterbi algorithm \cite{Durand.2004} then computes the most probable hidden variables $\argmax_{Z_{1:T}} \mathbb{P}(Z_{1:T} \mid X_{1:T})$.

\subsection{Our algorithm: TIMELY}
TIMELY combines pseudotime inference methods with inhomogenenous HMTs. The pseudotime inference algorithm establishes an intrinsic ordering of the cells based on morphology, and the HMT then finds inconsistent labels and proposes correct labels of the cells corresponding to the true cell types.

The input of TIMELY is a set of images together with noisy expert labels.
First, we use a convolutional neural network to learn meaningful feature representations of the cell images that are consistent with the morphology of the cells.
The convolutional network consists of three convolutional layers with $32$ filters each, where the filter size is $3 \times 3$. After each convolutional layer, there is a max-pooling layer with pooling size $2 \times 2$. A bottleneck of $50$ units, which provide the resulting feature vectors, is followed by two dense layers with $30$ hidden units each and an output layer. 
As an alternative, we also explored unsupervised methods such as autoencoders to learn feature representations of the images so that the training is not affected by noisy labels; this yielded qualitatively similar findings (results not shown). 

Next, a suitable pseudotime inference method is applied to calculate the pseudotimes, and the cells are ordered increasingly according to the pseudotime. We use either SCORPIUS or STREAM, depending on the topology of the data.
The sorted expert labels serve as the observed information in the HMT, and the hidden labels are the true cell types to be determined. We can use our background information about the dataset to fix the start probabilities $\bm{\pi}$ and the emission matrix $\bm{B}$, while the parameters of the transition matrices are learned by the generalized EM algorithm. Through the generalized Viterbi algorithm, we find the most probable true labels and the estimated cell type borders, which are unique due to the Markov assumption~\cite{abkowitz1996evidence}. 

\begin{figure}[h]
\begin{center}
\includegraphics[width=0.95\linewidth]{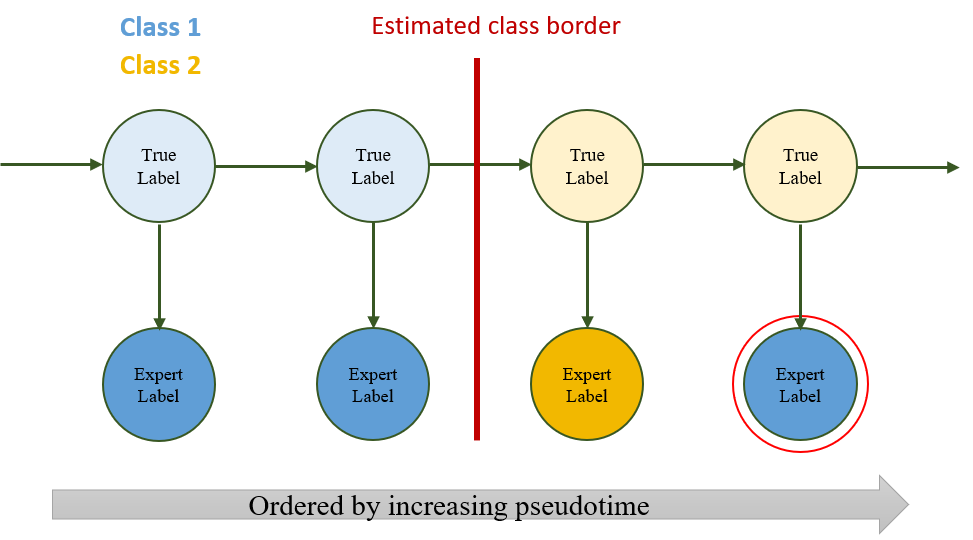} 
\end{center}
\caption{Inconsistent label.  The transition borders between the classes are derived through the generalized Viterbi algorithm. An image classification where the true label and the expert label do not coincide may be an error of the expert.}
\label{fig:wrong_label}
\end{figure}

Any inconsistencies between the true labels and the expert labels are potential mistakes by the expert (Figure \ref{fig:wrong_label}). Hematologists can reconsider the affected images and, if necessary, correct the labels of the cells. 
The method is summarized in Algorithm \ref{alg:method}. We implemented TIMELY in Python, and the library SciPy is used for maximizing the objective function in the generalized EM algorithm\footnote{The source code for the implementation can be found on \\\url{https://github.com/liu-yushan/label-consistency.}}. 

\begin{algorithm}[h]
\caption{TIMELY}
\label{alg:method}
\textbf{Input:} Images and noisy expert labels.\\
\textbf{Output:} Images with inconsistent labels and proposed labels.
\begin{algorithmic}[1]
\State Use background information about the dataset to define the topology of the HMT, the start probabilities $\bm{\pi}$, and the emission matrix $\bm{B}$.
\State Learn feature representations of the images using a convolutional neural network.
\State Choose a suitable pseudotime inference algorithm and calculate the pseudotimes of the feature vectors.
\State Sort the corresponding expert labels by increasing pseudotime.
\State Set up an HMT, where the sorted expert labels are the observed information.
\State Learn the parameters in the transition matrices $\bm{A^{(t)}}$ using the generalized EM algorithm.
\State Apply the generalized Viterbi algorithm to infer the most probable true labels.
\State Identify images with inconsistent labels by comparing the true labels with the expert labels.
\end{algorithmic}
\end{algorithm}

\section{COMPARISON TO BASELINES}

\subsection{Baseline methods}
We compare TIMELY to three baseline methods. As discussed in Section $1$, most algorithms are either robust to noise, find and remove noisy labels, or model label noise explicitly, but they do not propose new labels.

The algorithms $k$-nearest neighbors ($k$-NN) and $k$-nearest centroid neighbors ($k$-NCN) \cite{Sanchez.1997} find neighbors for each instance for a given distance measure. A commonly used distance measure for $k$-NN is the Euclidean distance, while for $k$-NCN, we add instances to the set of nearest neighbors for which the centroid of the new set is nearest to the considered instance. The label of the considered instance is then obtained by a majority vote. If the majority vote yields a different label than the initial label of the instance, or if there is a tie, the instance might be incorrectly labeled.
To compare our method with other methods that also propose corrections, we extend these two methods with generalized editing \cite{Koplowitz.1981}, i.\,e., we choose numbers $k$ and $k'$ with $(k+1)/2 \leq k' \leq k$ for $k$-NN and $k$-NCN. For each instance, if there are at least $k'$ nearest neighbors from a different cell type,  the cell type of the instance is changed to that type. Unlike in \cite{Koplowitz.1981}, we do not delete any samples. For both methods, we choose the numbers $k~=3$ and $k'=2$, which are common values in the literature \cite{Saez.2015, Sanchez.2003}.

We also compare TIMELY to cleanlab \cite{cleanlab, Northcutt.2019}, which is based on confident learning \cite{Northcutt.2017} and finds labeling errors. It estimates the noise rates by calculating the joint distribution between noisy and uncorrupted labels and then prunes inconsistent samples.

\subsection{Simulation data}
Since expert labels from real-world datasets are often noisy, we do not know the ground truth labels of the images. For comparing our algorithm to other methods in finding inconsistent labels, we simulate three datasets with different noise levels that mimic the cell differentiation setting. 
Each dataset consists of $250$ samples from five cell types, where the underlying topology is a chain. The process of simulating the datasets is the following:
\begin{enumerate}
    \item Sample $\bm{X} \in \mathbb{R}^{2 \times 250}$, where $\bm{X}$ is normally distributed.
    \item Sort the columns of $\bm{X}$ by increasing $\bm{X}_{1j}$, $1 \leq j \leq 250$.
    \item Define the corresponding ground truth labels $\bm{Y} \in \mathbb{R}^{250}$, where the entries $\bm{Y}_{50(i-1)+1:50i}$ are $i$ for $i \in \{1, \dots 5\}$.
    \item Apply mapping $\bm{P}$ to project $\bm{X}$ to a higher-dimensional space: $\bm{\Tilde{X}} = \bm{P}\bm{X} \in \mathbb{R}^{k \times 250}$. We choose $k=50$ to be consistent with our real-world datasets.
    \item Add noise level $l \in \{10, 20, 30\}$ to the ground truth labels $\bm{Y}$ by randomly changing $l\%$ of the entries in $\bm{Y}$ to different labels. The steps $1$ to $4$ are repeated for each noise level.
\end{enumerate}
The idea is that the samples have a low-dimensional ordering, corresponding to the pseudotemporal ordering, which can be retrieved by dimensionality reduction of the higher-dimensional feature vectors.

\subsection{Simulation results}
The results of the comparison is shown in Table \ref{tab:results}. The methods $k$-NN+edit and $k$-NCN+edit modify the labels during application, while $k$-NN, $k$-NCN, and cleanlab only find possible labeling errors. TIMELY finds labeling errors and proposes new labels without changing them directly. 

We compare the proposed labels with the ground truth labels to calculate the accuracy. The selected items are the instances that the algorithm marked as labeling errors.
While TIMELY finds errors in a magnitude that is similar to the noise level, the other methods mostly find too many errors, without increasing the recall. Only in one case, $k$-NCN has a higher recall than TIMELY. Our method has the highest accuracy, precision, recall, and $\mathrm{F}_1$ score in all the other cases.

Editing in $k$-NN and $k$-NCN often improves the $\mathrm{F}_1$ score compared to the versions without editing. However, edition of labels during application influences the classification of subsequent samples so that the accuracy drops if there are too many false positives.

\begin{table*}[h]
\begin{center}
{\caption{Comparison to baseline methods for simulation data. TIMELY outperforms all baseline methods in terms of accurately identifying and correcting noisy labels.}
\label{tab:results}}
\begin{tabular}{|c|c||c|c|c|c|c|c|}
\hline
Noise level & Metric & TIMELY & $k$-NN & $k$-NN+edit &$k$-NCN & $k$-NCN+edit & cleanlab\\
\hline
\multirow{5}{*}{10}   & Accuracy    &  $\bm{0.984}$ & -&$0.920$ & -& $0.944$ & -\\

&   Selected items &  $0.108$  &$0.164$& $0.152$& $0.152$&$0.144$ & $0.112$\\

 & Precision & $\bm{0.889}$& $0.561$&$0.579$ &$0.658$&$0.667$  &$0.679$ \\

 & Recall & $0.960$&$0.920$ &$0.880$ &$\bm{1.000}$ &$0.960$ &$0.760$ \\
 
 &   $\mathrm{F}_1$ score &  $\bm{0.923}$  &$0.697$ &$0.698$&$0.794$ &$0.787$ & $0.717$\\
 \hline
\multirow{5}{*}{20}   & Accuracy & $\bm{0.992}$ &-& $0.932$ & -& $0.820$ & - \\

&  Selected items & $0.192$ & $0.292$& $0.236$ &$0.324$ &$0.284$ &$0.256$\\

 & Precision & $\bm{1.000}$&$0.658$ & $0.797$ &$0.556$ &$0.634$ &$0.625$ \\

& Recall  & $\bm{0.960}$ &$ \bm{0.960}$ &$0.940$ &$0.900$ &$0.900$ &$0.800$\\

 &   $\mathrm{F}_1$ score &  $\bm{0.980}$  &$0.781$ &$0.863$&$0.687$ &$0.744$ & $0.702$\\
\hline
\multirow{5}{*}{30}  & Accuracy & $\bm{0.972}$& -&$0.792$ &-&$0.712$ &-\\

& Selected items &$0.300$ &$0.416$ &$0.340$ &$0.484$ &$0.404$ &$0.272$ \\

& Precision & $\bm{0.987}$&$0.673$ & $0.706$ &$0.537$ &$0.604$ &$0.809$ \\

& Recall &$\bm{0.987}$ &$0.933$ &$0.800$ &$0.867$ &$0.813$ &$0.733$\\

 &   $\mathrm{F}_1$ score &  $\bm{0.987}$  &$0.782$ & $0.750$& $0.663$ & $0.693$ & $0.769$\\
 \hline
\end{tabular}
\end{center}
\end{table*}

\section{APPLICATION TO REAL-WORLD DATASETS}
We apply TIMELY to two image datasets of stained white blood cells. All images were generated from a digital microscope (Cellavision, Siemens Healthineers AG) and labeled by an expert. Due to the challenges in manual labeling outlined in Section $1$, the labels are noisy and partly incorrect. For the preparation of the images, a thin blood film is wedged on a glass slide and stained. The digital microscope then locates the blood cells and creates corresponding images. Our datasets contain images from several patients. TIMELY is applied on the whole dataset to first find an ordering of all images, then it suggests a label for each image. For a new patient, images from the same developmental tree can be mapped onto the already calculated tree, and consistent labels can be read off the tree directly by making use of the already computed transition borders.

\subsection{Cell lineage line}

\begin{figure}[h]
\begin{center}
\includegraphics[width=0.48\linewidth]{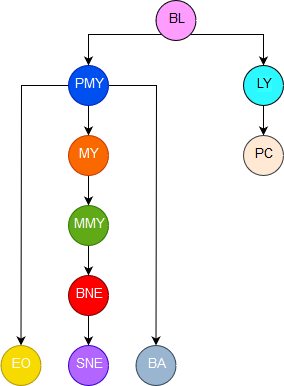} 
\end{center}
\caption{Blood cell lineage in the datasets. The two real-world datasets contain ten cell types.}
\label{fig:dataset_tree}
\end{figure}

\begin{figure}
{\centering
\begin{subfigure}[b]{0.09\textwidth}
\includegraphics[width=\textwidth]{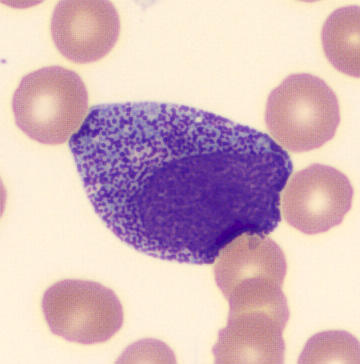}
\end{subfigure}
\begin{subfigure}[b]{0.09\textwidth}
\includegraphics[width=\textwidth]{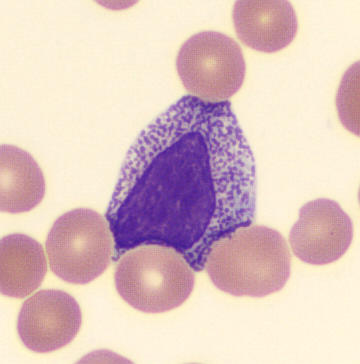}
\end{subfigure}
\begin{subfigure}[b]{0.09\textwidth}
\includegraphics[width=\textwidth]{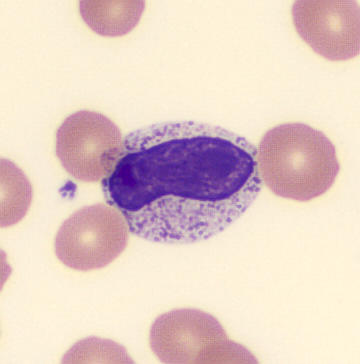}
\end{subfigure}
\begin{subfigure}[b]{0.09\textwidth}
\includegraphics[width=\textwidth]{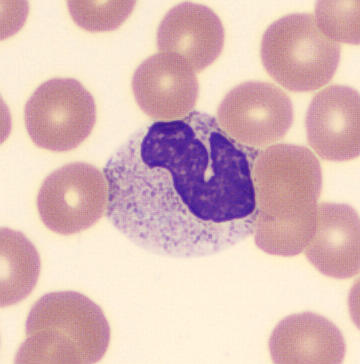}
\end{subfigure}
\begin{subfigure}[b]{0.09\textwidth}
\includegraphics[width=\textwidth]{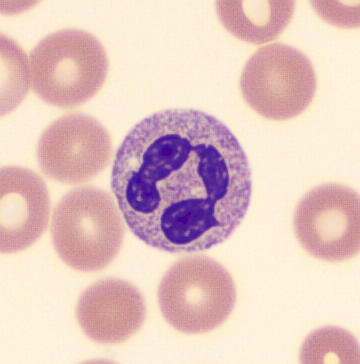}
\end{subfigure}
\caption{Example images. The images show the five cell types promyelocyte, myelocyte, metamyelocyte, band neutrophil, and segmented neutrophil.} 
\label{fig:dataset_images}}
\end{figure}

\paragraph{Dataset}
The first dataset consists of $1000$ cell images that contain five cell types of the development line granulopoiesis. The topology is a linear chain, and there are $200$ images labeled by an expert as belonging to each of the cell types promyelocyte (PMY), myelocyte (MY), metamyelocyte (MMY), band neutrophil (BNE), and segmented neutrophil (SNE). Figure \ref{fig:dataset_tree} shows the differentiation hierarchy; example images in this order are shown in Figure \ref{fig:dataset_images}.

\paragraph{Parameters in HMM}
We use our background knowledge about the dataset to fix the start probabilities $\bm{\pi}$ and the emission matrix $\bm{B}$. 
The dataset has five cell types, and we know that the root type in the development process is PMY. So we can fix the start probabilities to be
\begin{equation}
\bm \pi := \begin{pmatrix} 0.9 & 0.025 & 0.025 & 0.025 & 0.025\end{pmatrix}^T.
\label{eqn:start_prob}
\end{equation}
The first cell should be in the first cell type with high probability and in the other cell types with low probability.

The constant emission matrix $\bm{B}$ is based on estimations of an expert who could realistically estimate the probability of labeling errors. The emission matrix for the first dataset is shown in \eqref{eqn:expert_emission}. The more mature cell types band neutrophil and segmented neutrophil are fairly easy for humans to differentiate, while the first three cell types, especially myelocytes, are more difficult to label.

{
\begin{equation}
\begin{blockarray}{cccccc}
& \mathrm{PMY} & \mathrm{MY} & \mathrm{MMY} & \mathrm{BNE} & \mathrm{SNE} \\
\begin{block}{c(ccccc)}
\mathrm{PMY} & 0.7 & 0.25 & 0.04 & 0.005 & 0.005 \\
\mathrm{MY} & 0.23 & 0.52 & 0.24 & 0.005 & 0.005 \\
\mathrm{MMY} & 0.03 & 0.17 & 0.75 & 0.045 & 0.005 \\
\mathrm{BNE} & 0.005 & 0.005 & 0.03 & 0.82 & 0.14 \\
\mathrm{SNE} & 0.005 & 0.005 & 0.005 & 0.065 & 0.92 \\
\end{block}
\end{blockarray}
\label{eqn:expert_emission}
\end{equation}}

\paragraph{Pseudotime inference}
We use the algorithm SCORPIUS to compute the pseudotimes. Before SCORPIUS is applied, we use diffusion maps \cite{Coifman.2006, Haghverdi.2015} for dimensionality reduction,  after which SCORPIUS directly infers the trajectory without performing MDS.

\begin{figure*}[h]
{\centering
\begin{subfigure}[b]{0.27\textwidth}
\includegraphics[width=\textwidth]{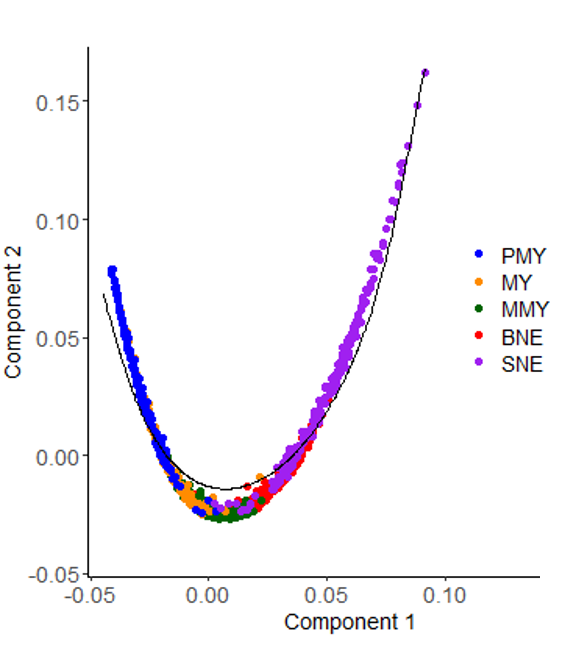}
\caption{}
\end{subfigure}
\hspace{4mm}
\begin{subfigure}[b]{0.25\textwidth}
\includegraphics[width=\textwidth]{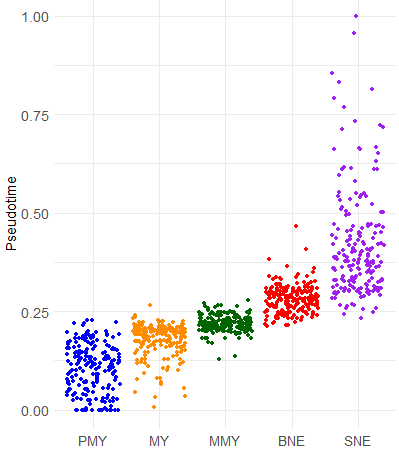}
\caption{}
\end{subfigure}
\hspace{5mm}
\begin{subfigure}[b]{0.38\textwidth}
\includegraphics[width=\textwidth]{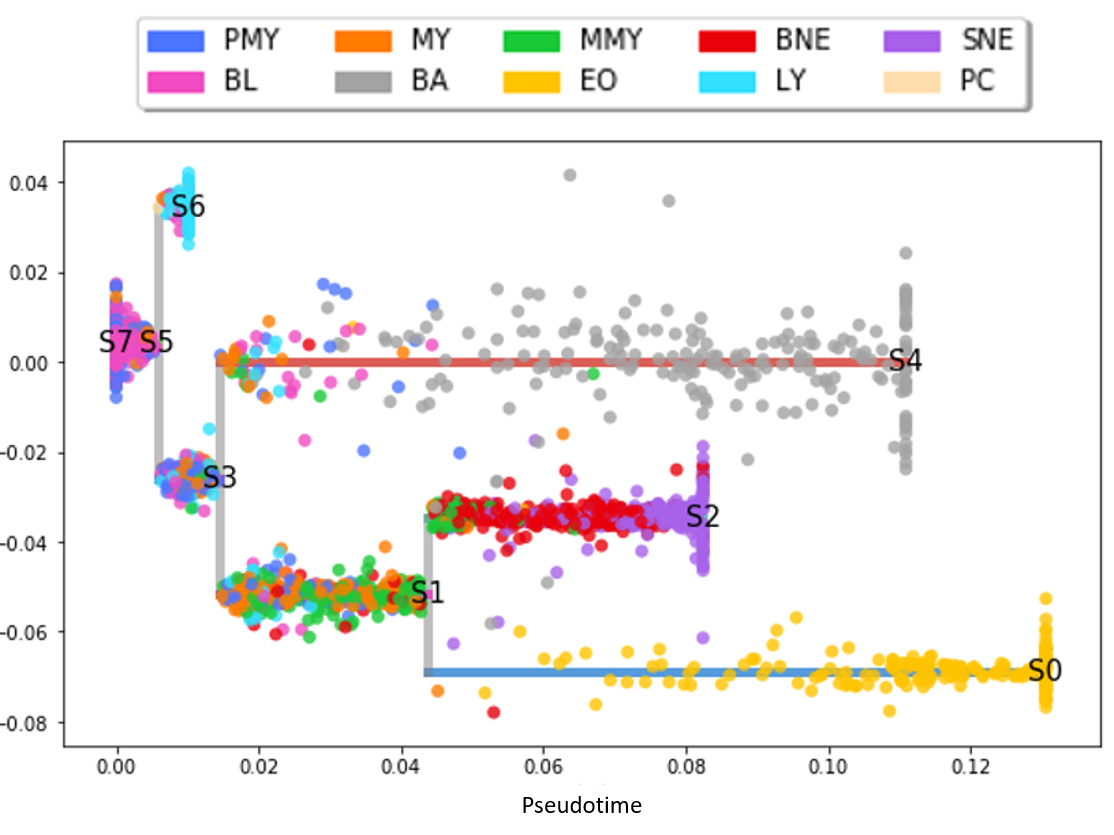}
\caption{}
\end{subfigure}
\caption{Pseudotime inference. (a) Diffusion maps are used for dimensionality reduction of the feature vectors. The trajectory (black line) is then inferred using SCORPIUS. (b) The cells are sorted by cell type. The pseudotimes, which are computed from the trajectory in (a), correspond to the cells' development progress. (c) STREAM calculates the pseudotimes of the cells, determines the branching points, and draws the corresponding tree.}
\label{fig:scorpius_pseudotime}}
\end{figure*}

Figure \ref{fig:scorpius_pseudotime}(a) shows the trajectory (black line) that is inferred by SCORPIUS for the cell development line. Each cell is represented by a point in the plot, which is colored according to its expert label. The pseudotime is normalized to be between $0$ and $1$. The value $0$ stands for the beginning of the development process, while the value $1$ represents the end of the development process. In Figure \ref{fig:scorpius_pseudotime}(b), the cells are grouped by cell type, and we see that with the development of the cells, the pseudotime increases. For myelocytes, for example, there are some cells with smaller pseudotimes; these might be instances with wrong labels. 

\begin{figure}[h]
\begin{center}
\includegraphics[width=0.95\linewidth]{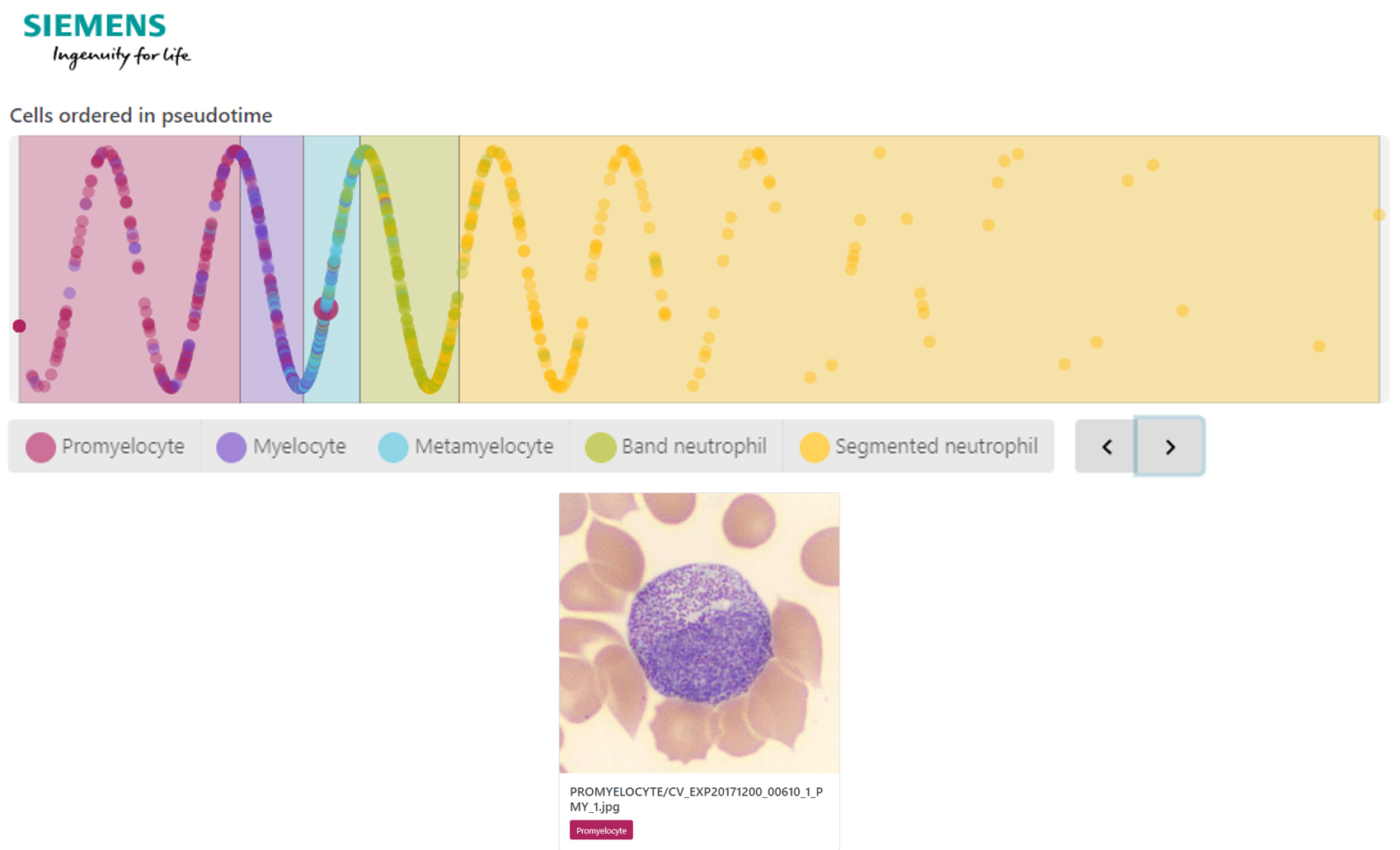} 
\end{center}
\caption{Visualization tool for experts. The cells are sorted by increasing pseudotime, and each cell type is differently colored. Inconsistent classifications can be identified by mismatching colors.}
\label{fig:tool}
\end{figure}

\paragraph{Visualization tool} 
After parameter optimization, the HMM finds unique transition borders between the cell types. We provide a visualization tool for viewing the images, which is shown in Figure \ref{fig:tool}. The images are ordered by the inferred pseudotime. Each image corresponds to one point, which is highlighted in the color corresponding to the expert label. The inferred transition borders from the HMM are integrated, and the spaces between the borders are colored according to the proposed cell types. Inconsistent classifications can be identified by mismatching colors. The expert can click on each point to see the corresponding image and navigate to neighboring cells by clicking on the arrows, thereby getting an intuition why a specific cell was marked as having an inconsistent label. 

\paragraph{Inconsistent labels}
The amount of consistent labels, where the hidden labels and expert labels coincide, is according to the HMM $72\%$, which means that there are $280$ images with potentially wrong labels.

\begin{figure}[h]
\begin{center}
\includegraphics[width=0.75\linewidth]{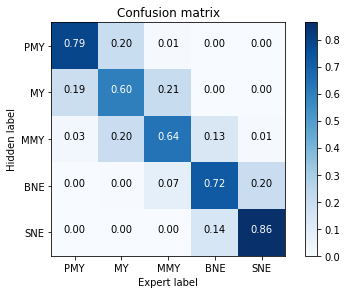} 
\end{center}
\caption{Confusion matrix for the first dataset. The confusion matrix shows some similarities to the emission matrix \eqref{eqn:expert_emission}. Most inconsistencies are in the cell types myelocyte and metamyelocyte.}
\label{fig:confusion_matrix}
\end{figure}

The confusion matrix in Figure \ref{fig:confusion_matrix} shows that the consistency for myelocytes and metamyelocytes is particularly low. Overall, the tendency of the values are similar to the expert's estimation of the emission matrix \eqref{eqn:expert_emission}. Experiments showed that the results are quite robust with respect to the emission matrix so that small changes in the estimations will not affect the results significantly.

The $280$ inconsistent images were given to a domain expert for reclassification. For $128$ of these images ($ 45.7\%$), the expert confirmed the previous labels. For the remaining $152$ cells, the expert either relabeled them as the cell types the HMM proposed, or she could not give a label with high confidence, which means that up to $54.3\%$ of the inconsistent images might have wrong labels. Most of the reclassifications affect the first three cell types in the development line, where the changes in the morphology can be very subtle.

\subsection{Cell lineage tree}

\paragraph{Dataset}
The second dataset consists of $1821$ cell images from ten classes that are part of a development process with branching points.
There are $200$ images labeled by an expert as belonging to each of the cell types promyelocyte (PMY), myelocyte (MY), metamyelocyte (MMY), band neutrophil (BNE), segmented neutrophil (SNE), blast (BL), basophil (BA), eosinophil (EO), and lymphocyte (LY). For the last class plasma cell (PC), there are only $21$ images.
Figure \ref{fig:dataset_tree} shows the underlying lineage tree of the dataset, where we have four end stages with no further transitions. Eosinophils and basophils also have the precursors myelocyte, metamyelocyte, and band neutrophil. However, these would have different staining colors as the precursors of the segmented neutrophil. Because those cell types are quite rare in the blood, they are not included in the dataset.

\paragraph{Parameters in HMT}
As we can see in Figure \ref{fig:dataset_tree}, the root cell should be a blast cell so that the entry for blast is very high in $\bm{\pi}$. The constant emission matrix $\bm{B}$ is again based on discussions with an expert and is a consistent extension of \eqref{eqn:expert_emission}. The five additional cell types should not be too difficult to differentiate from the cell types of the first dataset because they are part of different development lines. Only the blasts have some similarities to the promyelocytes, which are descendants of the blasts. The end stages segmented neutrophil, basophil, and eosinophil should be easy for experts to classify.

\paragraph{Pseudotime inference}
We apply the algorithm STREAM for inferring a reasonable tree for the dataset. Two of the three branching points in Figure \ref{fig:scorpius_pseudotime}(c) match the branching points from the cell lineage tree in Figure \ref{fig:dataset_tree}. However, the last one where the eosinophils branch off from the metamyelocytes is different. Generally, the eosinophils are far away from the other cell types in the feature space after dimensionality reduction. The connection point to the remaining tree might not be so accurate. Another reason could be that the precursor cells of segmented neutrophils and eosinophils look alike. Eosinophils have the same progenitor stages as the neutrophils that are only stained in a different color. The algorithm might identify the metamyelocyte also as a previous development stage of the eosinophil. The range of the pseudotimes are still plausible for all cell types.

\paragraph{Inconsistent labels}
The percentage of consistent labels according to the HMT is $69\%$, which means that there are $564$ images with potentially wrong labels. The blasts and promyelocytes seem to be mixed up often, while basophils and eosinophils have high agreement between hidden labels and expert labels, probably because of their distinct staining colors. The agreement for lymphocytes is also very high since cells from different development lines are usually easier to differentiate. 

\begin{figure}[H]
\begin{center}
\begin{subfigure}[b]{0.09\textwidth}
\includegraphics[width=\textwidth]{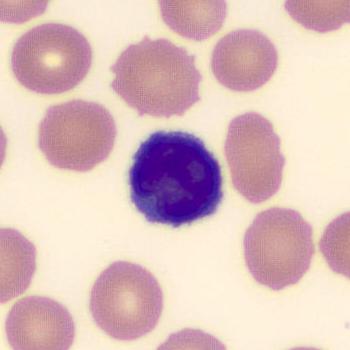}
\caption{}
\end{subfigure}
\begin{subfigure}[b]{0.09\textwidth}
\includegraphics[width=\textwidth]{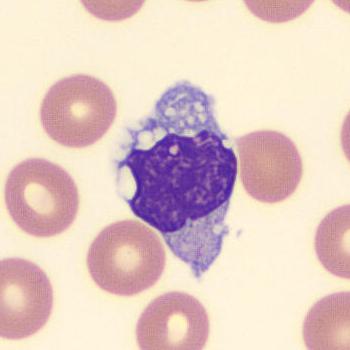}
\caption{}
\end{subfigure}
\begin{subfigure}[b]{0.09\textwidth}
\includegraphics[width=\textwidth]{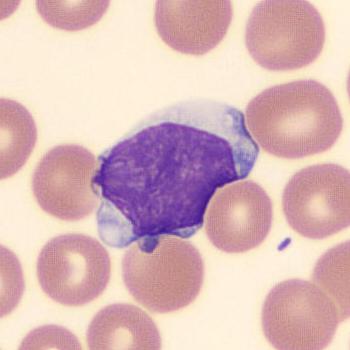}
\caption{}
\end{subfigure}
\begin{subfigure}[b]{0.09\textwidth}
\includegraphics[width=\textwidth]{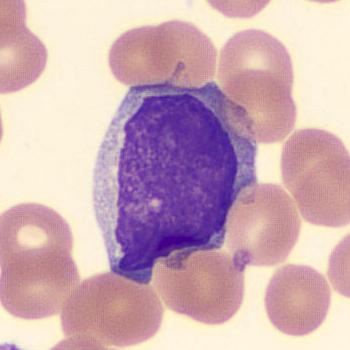}
\caption{}
\end{subfigure}
\begin{subfigure}[b]{0.09\textwidth}
\includegraphics[width=\textwidth]{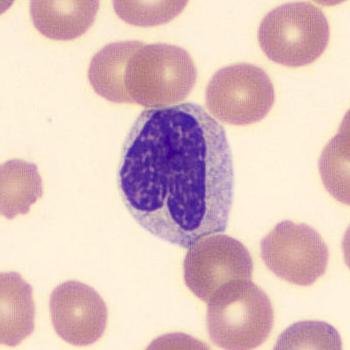}
\caption{}
\end{subfigure}
\caption{Examples of reclassifications. (a) BA $\rightarrow$ LY. (b) PC~$\rightarrow$~LY. (c) PC $\rightarrow$ BL. (d) PMY $\rightarrow$ BL. (e) MMY~$\rightarrow$~BNE.} 
\label{fig:reclass_hmt}
\end{center}
\end{figure}

The $564$ inconsistent images were given to a domain expert for reclassification. The expert confirmed for $341$ images their previous labels so that up to $40.1\%$ of the inconsistent images might have wrong labels. Most reclassifications affect promyelocytes, myelocytes, and metamyelocytes, the first three cell types in the granulopoiesis line. The cells are mostly reclassified as the progenitor of the cell type that the experts have determined. Figure \ref{fig:reclass_hmt} shows five example images that the expert has reclassified with the proposed labels from the HMT.

\section{DISCUSSION}
We have introduced TIMELY, a human-centred approach for increasing labeling consistency in medical imaging for cell type classification. TIMELY takes as input cell microscopy images along with noisy expert labels, identifies inconsistent labels and suggests alternative, consistent labels based on  a two-step procedure. In the first step, TIMELY establishes an intrinsic order between cells using a pseudotime inference algorithm. In the second step, TIMELY builds a Markov model upon the ordered cells and their noisy labels. Depending on the complexity of the dataset's topology, an HMM or an HMT is used. 

We combine pseudotime estimations with interpretable HMTs to establish a 
system that assists the annotating hematologist to generate more consistent cell classifications. By sorting the cells according to the pseudotime, we enable the hematologist to consider each cell in a neighborhood of cells that have a similar morphology, thereby assisting him in making consistent decisions (Figure \ref{fig:tool}). In addition, we transparently and explicitly encode domain knowledge in form of differentiation hierarchies (Figure \ref{fig:dataset_tree}), start probabilities \eqref{eqn:start_prob}, and an expert-driven emission matrix \eqref{eqn:expert_emission}, reflecting prior experience on the likelihood of labeling errors. Taken together, this allows the hematologist to develop an intuitive understanding on why specific cells are suggested as being inconsistently labeled and helps a more readily adoption in practice.

Manually labeling cells is also a time-consuming process, and our method can be applied to reduce the time experts spend on this task. Thus, once parameters of an HMT are optimized, new images from the same developmental tree can be mapped onto the already calculated tree, and consistent labels can be read off the tree directly by making use of the already computed transition borders.

Some modern digital microscopes have a functionality that automatically suggests labels for cell images. An additional use case of TIMELY is the application to such automatically generated labels since they are often noisy, and in addition, the classification algorithm does not include all possible cell types. These labels would then serve as the observed information of the HMT, and only the inconsistent labels will be given to the expert for reclassification. Such a system would be a further step towards an automated machine learning method that supports humans in a meaningful way.

\section{CONCLUSION}
TIMELY is a probabilistic approach for improving the cell labelings of experts that combines pseudotime inference algorithms and hidden Markov trees. Using pseudotime for ordering cells intrinsically leads to labels that are consistent with the morphology of the cells. We incorporate necessary background information about the data into an inhomogeneous hidden Markov tree, which makes use of the pseudotemporal ordering. Our model does not only find noisy labels like other filtering methods, but it also suggests alternative, consistent labels. It is able to identify and correct noisy labels with higher accuracy and precision than state-of-the-art methods for identifying noisy labels. Application on two real-world datasets and the subsequent reclassification by an expert demonstrate that the labelings have indeed been improved by our algorithm.

\bibliography{ecai.bib}
\end{document}